\documentclass[11pt,a4paper]{article}
\usepackage[hyperref]{acl2020}
\usepackage{standalone}
\usepackage{mystuff-nlp}
\usepackage{times}
\usepackage{tikz}
\usepackage{url}
\usepackage{latexsym}
\usepackage{color}
\usepackage{float}
\usepackage{amsfonts,amssymb,amsmath,bm,bbm}
\usepackage{algpseudocode}
\usepackage{graphicx,subfigure}
\usepackage{booktabs}
\usepackage{multirow}
\usepackage{verbatim}
\usepackage{enumitem,array}
\usepackage{setspace}

\usepackage[titlenumbered,ruled]{algorithm2e}

\usepackage{microtype}

\newcommand{\example}[1]{`#1'}
\newcommand{\sysname}{\textsc{Mardi}}
\aclfinalcopy % Uncomment this line for the final submission
\def\aclpaperid{2702} %  Enter the acl Paper ID here

\title{One Model to Recognize Them All:\\ Marginal Distillation from NER Models with Different Tag Sets}

\author{Keunwoo Peter Yu \and Yi Yang\\
	ASAPP Inc.\\
	New York, NY 10007\\
	{\tt \{yyang+peter\}@asapp.com}
        }

\date{}

\begin{document}
\maketitle
\begin{abstract}
Named entity recognition (NER) is a fundamental component in the modern language understanding pipeline. Public NER resources such as annotated data and model services are available in many domains. However, given a particular downstream application, there is often no single NER resource that supports all the desired entity types, so users must leverage multiple resources with different tag sets. This paper presents a marginal distillation (\sysname) approach for training a unified NER model from resources with disjoint or heterogeneous tag sets. In contrast to recent works, \sysname~merely requires access to pre-trained models rather than the original training datasets. This flexibility makes it easier to work with sensitive domains like healthcare and finance. Furthermore, our approach is general enough to integrate with different NER architectures, including local models (e.g., BiLSTM) and global models (e.g., CRF). Experiments on two benchmark datasets show that \sysname~performs on par with a strong marginal CRF baseline, while being more flexible in the form of required NER resources. \sysname~also sets a new state of the art on the progressive NER task.%\sysname~significantly outperforms the start-of-the-art model on the task of progressive NER.
\end{abstract}

% ***********************************************************
% Introduction
\section{Introduction}
\label{sec:intro}

% 1. Most of the NER datasets are incomplete and different use cases require multiple datasets. a) biomedical-spec NER datasets - also want to leverage datasets news domain; b) extend tag-set (new requirements).
Named Entity Recognition (NER) is the task of locating and categorizing spans of text into a closed set of classes, such as people, organizations, and locations. As a core information extraction task, NER plays a critical role in many language processing pipelines, underlying a variety of downstream applications including relation extraction~\cite{mintz2009distant} and question answering~\cite{yih2015semantic}.  Although many NER datasets have been created for various domains, a practical obstacle for applying NER to a specific downstream application is that there is often a mismatch between the application-desired entity types and those supported by a single tag set. The most common scenario is selective annotation that has been discussed by~\newcite{beryozkin2019joint} and~\newcite{greenberg2018marginal}. As annotating entities is expensive, collecting NER annotations for a particular domain usually focuses on domain-spec entity types and relies on existing datasets to identify entities of general types. Therefore, we have to consolidate NER corpora with multiple tag sets to build a joint named entity recognizer that covers the union of entity tags.

In addition, a substantial amount of NER resources exist in the form of models or services where the original data annotations are not available. This circumstance arises when the source domain is sensitive, such as medicine and finance, so only very limited access to data is allowed for privacy and security reasons. Another common case is when a production NER system has undergone multiple rounds of training and retraining on data sampled from different time periods, and it is impossible to infer the exact data set that contributed to the complicated system.  A special case of the problem, progressive NER, has been studied by~\newcite{chen2019transfer}, in which they adapt a source model using the target data with new tag categories appear, without accessing to the source data. The setting poses additional challenges for training the unified NER model to recognize entities defined by all tag sets.

\iffalse
One important challenge of sequence labeling tasks concerns dealing with the change of the application domain
during time. For example, in standard concept segmentation
and labeling (Wang, Deng, and Acero 2005), semantic categories, e.g., departure or arrival cities, vary according to
new scenarios, e.g., low-cost flight or budget terminal were
not available when the Automatic Terminal Information Service (ATIS) corpus was compiled (Hemphill, Godfrey, and
Doddington 1990). Similar rationale applies to another very
important sequence labeling task, Named Entity Recognition (NER), where entities in a domain are continuously
evolving, e.g., see the smart phone domain.

Our transfer
learning (TL) techniques enable to adapt the source model
using the target data and new categories, without accessing to the source data.

We train an initial model on source data and transfer it to a model that
can recognize new NE categories in the target data during a subsequent step, when the source
data is no longer available.
\fi

% 3. In this work, we aim at training a unified NER model that can recognize entities of types that are supported by either separate models or datasets with different tag sets. The tag sets can be disjoint or heterogeneous where the semantic meanings of tags of two sets overlap (e.g., Name of set 1 and First Name and Last Name of set 2).  Most prior works assume access all the training datasets.  Moreover, they focus on CRF-based models and lack of generality.
To overcome the aforementioned difficulties, we introduce a novel task of training a unified NER model using pre-trained models with different tag sets. In particular, the tag sets can be disjoint~\cite{greenberg2018marginal} or heterogeneous~\cite{beryozkin2019joint} where we can induce a hierarchy from the tag sets (e.g., \texttt{First Name} and \texttt{Last Name} are children of \texttt{Name}). A na\"ive approach would infer the proper tag sequence using the predictions from multiple tag sets through a post-processing step. However, the post-processing requires heuristics --- such as choosing the tag with the highest probability score --- to consolidate conflicted tag sequences, which often works poorly in practice. As the individual models are estimated on datasets whose scales may be remarkably inequal, probability scores of different models are generally not comparable.

% 4. To this end, we propose a general and flexible framework based on knowledge distillation technology.  Models trained on different tag sets are teachers and distill to a unified student model. Easily fit disjoint setting, heterogeneous tag setting and progressive NER setting. We propose a simple way to distill CRF models.
To this end, we present marginal distillation (\sysname), a simple yet flexible framework inspired by the knowledge distillation technology~\cite{hinton2015distilling}. \sysname~distills the knowledges of pre-trained NER models with different tag sets into a unified model without accessing the training data. The setting diverges from the typical application of knowledge distillation, model compression, in which a small model (i.e., student) is trained to mimic a pre-trained, larger model (i.e., teacher). It resembles the specialist-generalist distillation setup presented in~\cite{hinton2015distilling}, where a generalist model is trained to distill knowledge from an ensemble of specialist models. However, a specialist model concerns a subset of classes in the generalist label set, while the relationships between tags of different label sets are often beyond simple one-to-one mappings. Furthermore, as NER is a sequence tagging problem, it is unclear how to perform distillation for structured prediction models like conditional random fields (CRF). %\sysname~addresses the challenges by carrying out two levels of marginalization.
To address the challenges, we first construct a tag hierarchy~\cite{beryozkin2019joint} to align tags of multiple label sets. The probabilities of the child nodes are summed together to match the probability of a parent node in the hierarchy. Then, we propose a simple but effective strategy to distill knowledge from a CRF model, where the distillation target is the token-level marginal distribution that can be computed using dynamic programming.

%The key distinctions are three-fold: a) manipulation of a single dataset is required to create an ensemble, while the NER datasets with multiple tag sets exist everywhere; b) each specialist model concerns a subset of classes in the generalist label set, while entity types in two tag sets constitute relationships that are often beyond simple one-to-one mappings; c)

% Compared to competitive systems, a) does not require to access data, but also can work with data if they are available;
% b) general framework, not limited to the model architecture, local and global is okay.
% To show it works with global model, we further propose a simple way to distill CRFs, for the first time

% 5. Contributions include experiment results

%i) we introduce a novel task, UmNER, to integrate heterogeneous NER resources; ii) we present \sysname~to tackle UmNER, which transfer knowledge from pre-trained NER models to a unified NER model; iii) we propose a simple yet effective method for CRF knowledge distillation using the node marginals; iv) in experiments on standard benchmark datasets, we show that \sysname~achieves similar results to the strong marginal CRF method even without accessing the annotations. \sysname~significantly outperforms the previous state-of-the-art method on the progressive NER task.
Our contributions are:
\begin{itemize}
\item We introduce a novel task to integrate heterogeneous NER resources.
\item We present \sysname~to tackle the novel task, which transfers knowledge from pre-trained NER models to a unified NER model.
\item We propose an effective method for CRF knowledge distillation using node marginals.
\item Experiments on standard datasets show that \sysname~achieves state-of-the-art results on consolidating NER systems with multiple tag sets without accessing raw annotations.
\end{itemize}

% ***********************************************************
% Background
\section{Background}
\label{sec:background}

In this section, we first introduce a standard neural architecture for NER that this work builds upon and then summarize previous work on joint training a NER model on datasets with multiple tag sets. Finally, we briefly describe knowledge distillation and its application to model compression.

\subsection{Neural networks for NER}
% BiLSTM-CRF
In recent years, neural network architectures such as convolutional neural networks (CNN) and recurrent neural networks (RNN) have revolutionized the field of natural language processing. \newcite{lample2016neural} introduce the bi-directional long short-term memory networks with conditional random fields (BiLSTM-CRF) that has become the standard architecture for NER.

\begin{figure}[t]
\centering
\includegraphics[scale=.44]{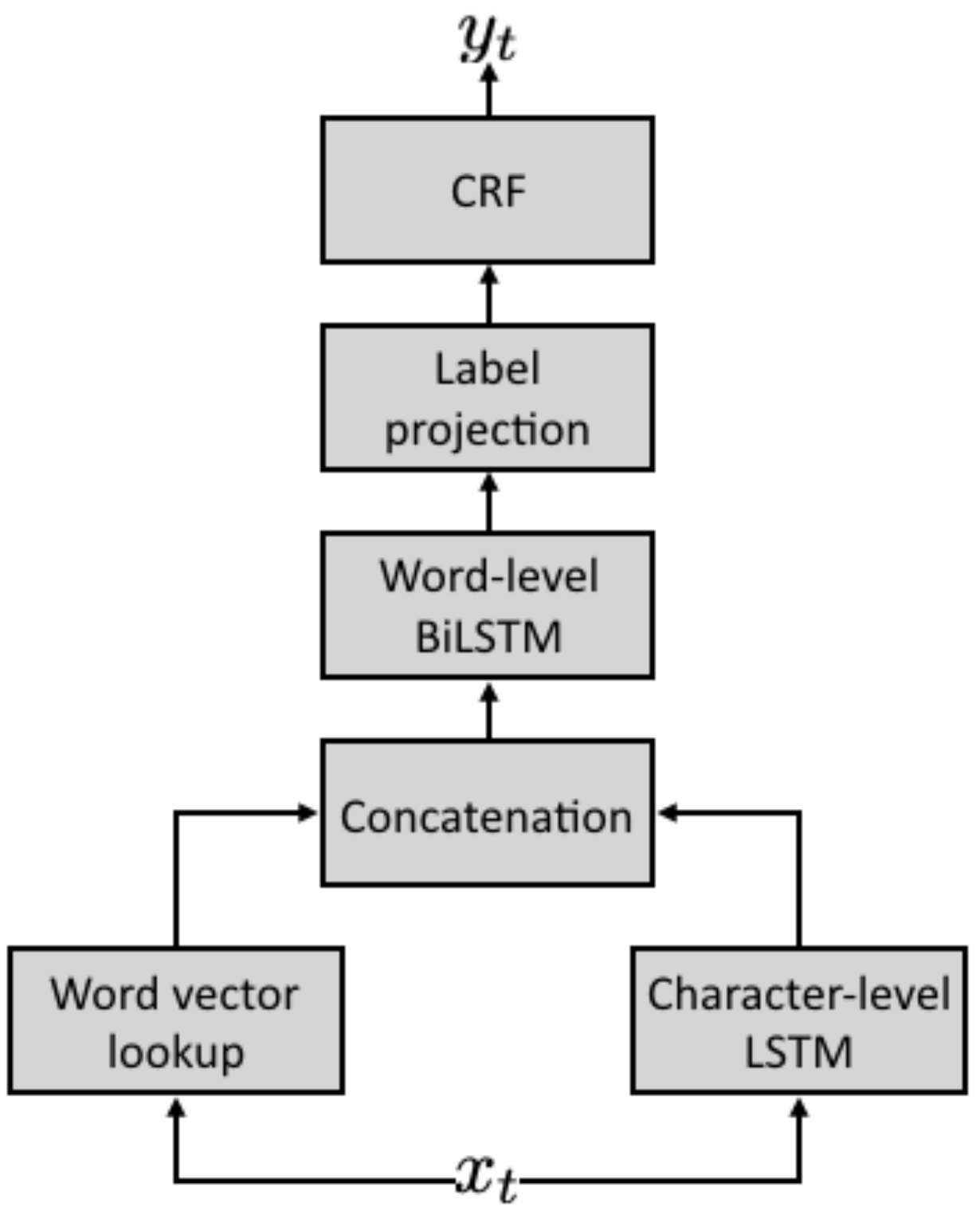}
\caption{The neural NER architecture~\cite{lample2016neural} that underlies most of the models studied in this work.}
\label{fig:bilstm-crf}
\end{figure}

As depicted in \autoref{fig:bilstm-crf}, given a sequence of tokens $\{ x_t \}_1^T$, BiLSTM-CRF first encodes each token into a vector representation using a character-level LSTM model. We adopt a uni-directional character-level LSTM model, as we found it performs slightly better than the BiLSTM model. The vector is then concatenated with a word embedding vector\footnote{We employ the publicly available 100-dimensional GloVe vectors~\cite{pennington2014glove} to initialize the word embeddings for all the models. } and the new vector is fed into a word-level BiLSTM model to produce a contextual representation for each token. This feature vector is then projected to the label dimension $L$ using a linear layer, representing the emission scores for predicting the token to the tags. Finally, BiLSTM-CRF incorporates the label-label transition scores with a conditional random field (CRF) layer~\cite{lafferty2001conditional}, to generate a globally optimized tag sequence prediction.

\subsection{Marginal CRF}
% Marginal CRF was proposed by ... to jointly train models.
In order to jointly train a NER model with multiple tag sets, \newcite{greenberg2018marginal} propose marginal CRF that is a variant of the BiLSTM-CRF model described above. Consider each dataset is partially annotated in which the annotations for the labels of other tag sets are unobserved. Specifically, a token that is labeled as \texttt{O} (i.e., not an entity) in a dataset can possibly be an entity token of another tag set.

\begin{figure}[t]
\centering
\includegraphics[scale=.4]{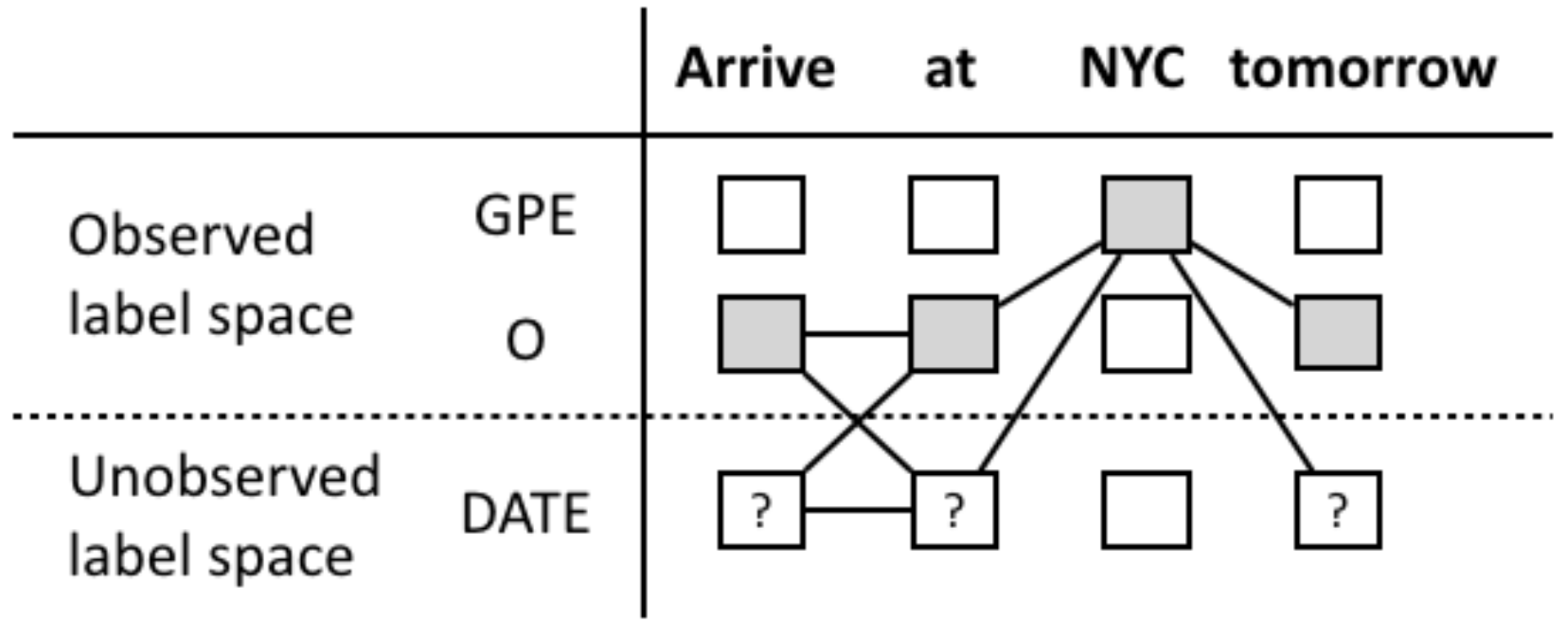}
\caption{Illustration of the marginal CRF method~\cite{greenberg2018marginal}. Two tag sets that one contains \texttt{GPE} and the other contains \texttt{DATE} are shown in the example. Here the annotations for the first tag set are given. Tokens labeled as \texttt{O} could potentially be either \texttt{O} or \texttt{DATE} of the second tag set. Marginal CRF marginalizes over all potential sequences.}
\label{fig:marginal-crf}
\end{figure}

To train a CRF model for the unified tag set, marginal CRF learns to score a partially observed tag sequence by marginalizing over unobserved paths. As shown in~\autoref{fig:marginal-crf}, a token labeled as  \texttt{O} potentially takes any entity type label from any of the other datasets. Thus, the probability of the partially observed tag sequence is actually the marginal probability of observed entities according to the unified tag set. The marginal can be calculated efficiently using the forward algorithm.

% The method was further extended by to
The method was further extended by~\newcite{beryozkin2019joint} to tackle label sets whose tags constitute relationships beyond simple one-to-one mappings. For example, the \texttt{GPE} label in the OntoNotes 5.0~\cite{weischedel2013ontonotes} tag set corresponds to three entity labels (\texttt{CITY}, \texttt{STATE}, and \texttt{COUNTRY}) defined in the I2B2 2014~\cite{stubbs2015annotating} tag set. To capture the semantic relationships between tags, a tag hierarchy in which \texttt{GPE} is the parent node of \texttt{CITY}, \texttt{STATE}, and \texttt{COUNTRY} is manually constructed. Given the tag hierarchy, the training procedure for the marginal CRF model is similar. The source probability of a sequence containing entities of a parent label type equals the target marginal probability over the child entity types.

% require data, only work with sequence model crf

\subsection{Knowledge distillation}
% knowledge distillation is model compression method in which a small model is trained to mimic a pre-trained, larger model (or ensemble of models) by ...
Knowledge distillation is first proposed by~\newcite{bucilua2006model} as a model compression method in which a small model is trained to mimic a pre-trained, larger model (or ensemble of models). It was generalized by~\newcite{hinton2015distilling} and now widely adopted in compressing deep neural networks with millions to billions of parameters into shallower networks with significantly smaller numbers of parameters. The pre-trained source model is typically referred to as the ``teacher'' and the target model is called the ``student''.

% Summary of the model
Concretely, we can transfer the knowledge in the teacher model to the student model by forcing them to have a similar prediction for any input instance.  This can be achieved by training the student model to minimize a loss function where the target is the distribution of class probabilities predicted by the teacher model. The typical choice of the loss function is the Kullback-Leibler (KL) divergence between the distributions, $D_{\text{KL}}(\mathbf{q} \parallel \mathbf{p})$, where $\mathbf{p}$ and $\mathbf{q}$ are the source and target output label distributions respectively. The distribution can be attained with a softmax function:
\[
p_i = \frac{\exp{(z_i / \tau)}}{\sum_j\exp{(z_j / \tau)}},
\]
where $z_i$ is the model logit for class $i$ and $\tau$ is a temperature parameter that controls the shape of the distribution for distilling richer knowledge from the teacher. We found $\tau=1$ works well for our application in practice, which is consistent with findings in some recent works~\cite{kim2016sequence}. The KL loss is also referred to as the ``distillation loss'' in literature.

% can use data annotations if available
In addition to the distillation loss, it is also beneficial to train the student model to predict the ground truth labels using the standard cross-entropy loss, dubbed as the ``student loss''. The overall objective is a linear combination of the distillation loss and the student loss. Note that the student loss is optional for knowledge distillation, which demonstrates the flexibility of the framework that makes it a perfect fit for the task of training a unifed NER model.

% ***********************************************************
% Model
\section{Marginal Distillation (\sysname)}
\label{sec:model}

We formally present marginal distillation (\sysname) in this section. Specifically, \sysname~is a general NER model unification framework that works with both disjoint tag sets and heterogeneous tag sets. We start with building a tag hierarchy to align tags of different label sets, and then show how to transfer knowledges from models pre-trained on different tag sets to a unified NER model with \sysname. Finally, we present a simple strategy to distill knowledge from a pre-trained CRF model to a student CRF model.

\subsection{Building a tag hierarchy}
Different tag sets often contain semantically inequivalent but related entity types. Downstream NER applications could be interested in types of entities of different granularity.  For example, \texttt{GPE} in the OntoNotes 5.0 tag set corresponds to three fine-grained tags,  \texttt{CITY},  \texttt{STATE}, and  \texttt{COUNTRY}, in the I2B2 2014 tag set. To address this phenomenon, \newcite{beryozkin2019joint} propose to build a tag hierarchy to unify multiple tag sets.

\begin{figure}[t]
\centering
\includegraphics[scale=.24]{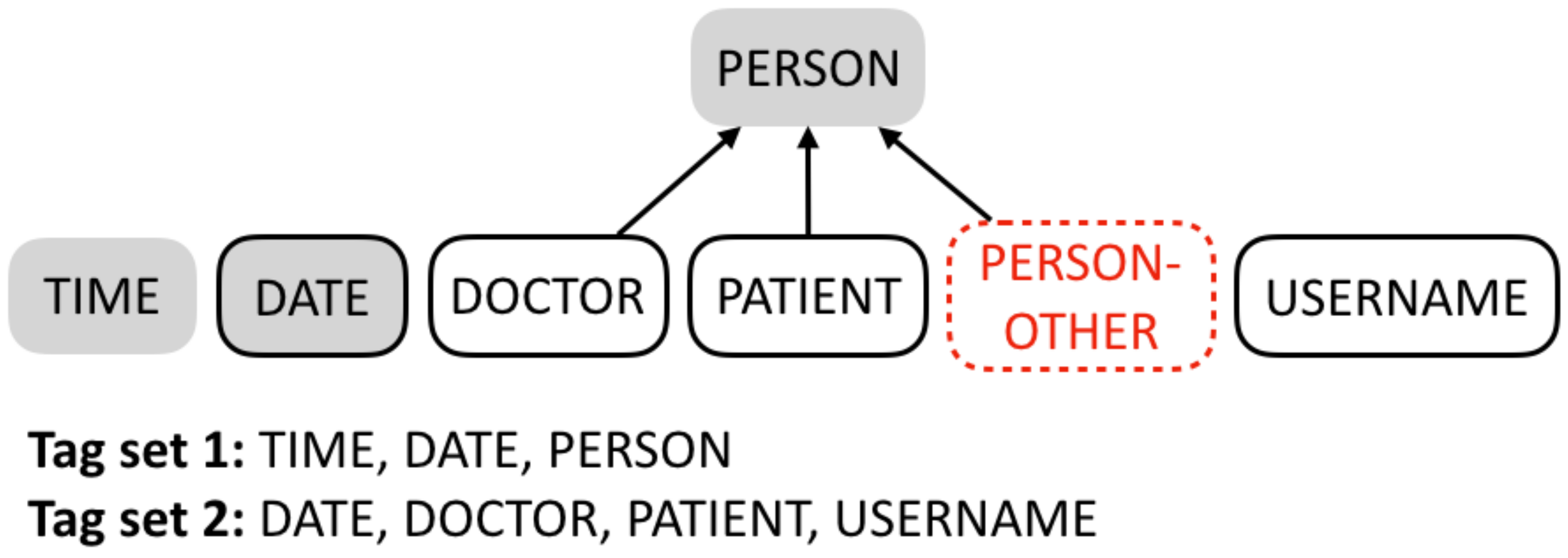}
\caption{A tag hierarchy constructed for two tag sets, whose tags are represented by filled nodes and bordered nodes respectively. The \texttt{PERSON-OTHER} tag is introduced to fill the remaining semantic space of the \texttt{PERSON} tag.}
\label{fig:tag-hierarchy}
\end{figure}

As displayed in~\autoref{fig:tag-hierarchy}, a tag hierarchy is a directed acyclic graph (DAG), in which each node represents a semantic tag of a label set (e.g.,  \texttt{GPE} and \texttt{CITY}).  A directed edge between the parent node $p$ and the child node $c$, $p \rightarrow c$, indicates that $c$ is a hyponym or finer-grained tag of $p$ and $c$ captures a subset of the semantics of $p$. As shown, we include three directed edges each of which is between \texttt{GPE} and one of \texttt{CITY},  \texttt{STATE}, and  \texttt{COUNTRY} to capture their semantic relationships. In many cases, we need to introduce an additional edge between the parent node and a placeholder child node (e.g., \texttt{PERSON-OTHER}), as the present child nodes fail to cover the full semantic space corresponding to the parent node. %Finally, a special \texttt{T-OTHER} tag is employed to fill the gap between the semantic spaces covered by the unified tag set and the individual tag set. It serves with a similar role as the dustbin label used by~\newcite{hinton2015distilling}, which takes care of label space outside of a specific tag set when we distill knowledge from a pre-trained model with the tag set.

One caveat about the tag hierarchy that \newcite{beryozkin2019joint} do not mention is that the hierarchy is not capable to align complicated tags whose semantics intersect but no specific hypernym-hyponym relationship exists. Such example tag pairs including (\texttt{CARDINAL}, \texttt{AGE}) and (\texttt{CARDINAL}, \texttt{STREET}) when aligning the tag sets of OntoNotes 5.0 and I2B2 2014. We leave the resolution of this issue for future work.

% We generally follow~\newcite{beryozkin2019joint} to manually construct the hierarchy, with the exception on how to handle the \texttt{OTHER} semantic for each tag set. As illustrated in~\autoref{}, we adopt a single \texttt{OTHER} tag for all the tag sets instead of using distinct tags for different tag sets, i.e., \texttt{T-OTHER}. Our \texttt{OTHER} tag serves with a similar role as the dustbin label used by~\newcite{hinton2015distilling}, which

\subsection{Marginal distillation over the hierarchy}
% setting
Given the hierarchy, our goal is to build an NER system that can extract any type of entities corresponding to a tag hierarchy node. To achieve this, we aim at training a NER model that predicts the most fine-grained tags in the hierarchy. The entity extraction results for an ancestor tag can be inferred by aggregating the extracted entities corresponding to its fine-grained descendants.

\begin{figure}[t]
\centering
\includegraphics[scale=.25]{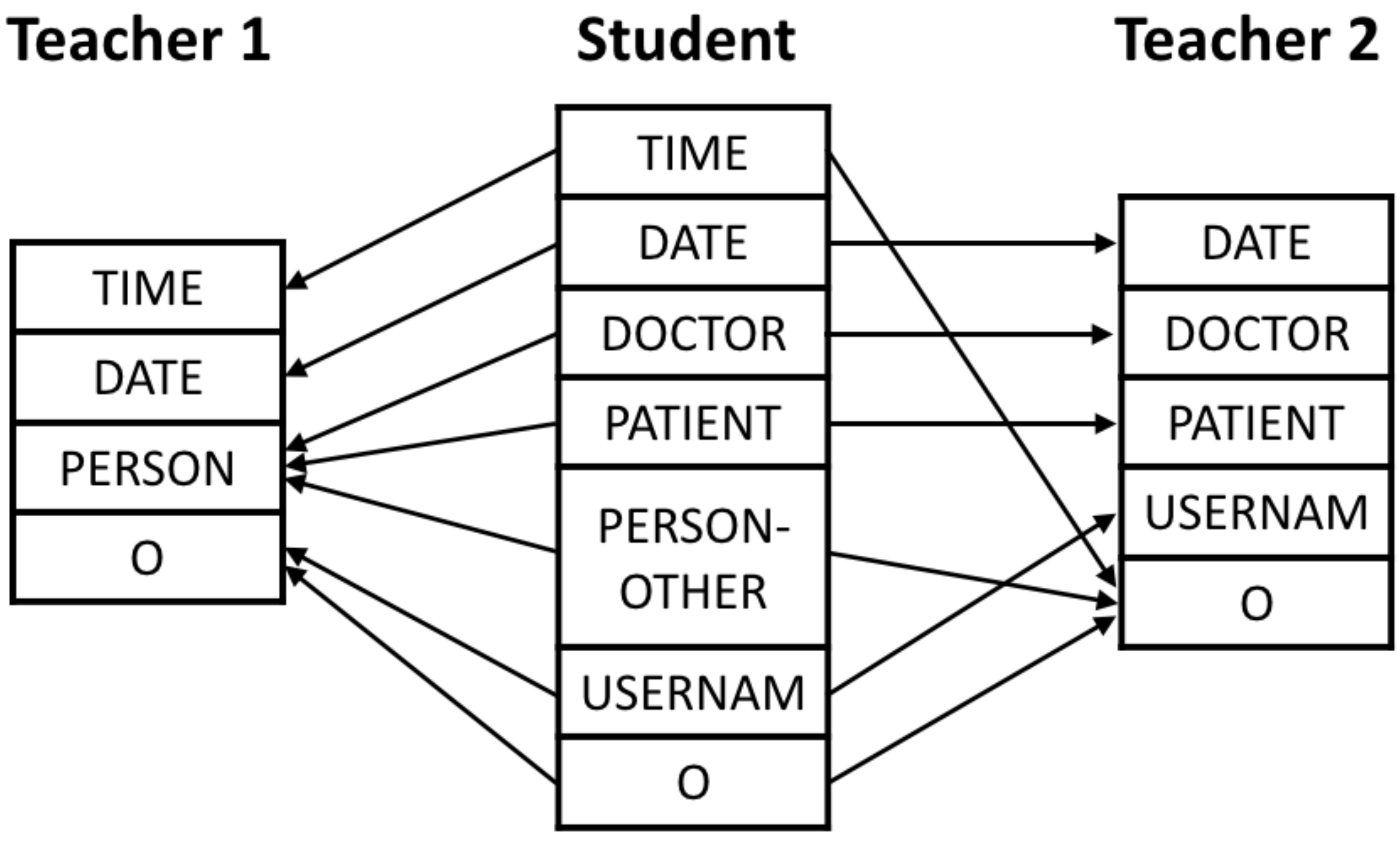}
\caption{Illustration of the proposed method, \sysname, to distill knowledge from two teachers pre-trained on datasets corresponding to the tag sets presented in~\autoref{fig:tag-hierarchy} to a student.}
\label{fig:mardi}
\end{figure}

% distillation loss between p and q
As shown in~\autoref{fig:mardi}, \sysname~transfers knowledge from multiple teacher models to a student NER model, where the student is trained to make predictions on the unified tag set that consists of all the fine-grained tags in the hierarchy. Particularly, given an input sequence $\mathbf{x} = \{ x_1, x_2, \cdots, x_T \}$, a distillation loss is computed between the predicted tag sequence of the student model and that of a teacher model $k$: %\footnote{The KL-divergence criterion used for training the student model is equivalent to minimizing the cross entropy of the soft target labels.}
\begin{equation}
\ell_k = - \sum_{t=1}^T \sum_{i=1}^{L_k} q_{t,i} \log p_{t,i},
\label{eq:loss}
\end{equation}
where $q_{t,i}$ is the soft target label for class $i$ of the $k$-th teacher's tag set, $L_k$ is the number of labels in the tag set, and $p_{t,i}$ can be obtained by summing the probabilities of $i$'s descendant fine-grained tags in the hierarchy,
\begin{equation}
p_{t,i} = \sum_{j \in \text{DescendantLeaf}(i)} p_{t,j},
\label{eq:marginal}
\end{equation}
where $p_{t,j}$ is the predicted probability from the student for the $j$-th class in the unified tag set.

% data distillation loss so overall loss
In addition to the distillation loss, to establish a fair comparison with prior work, we also consider the student loss for some of our experiments. The student loss replaces the soft target label $q_{t,i}$ with the ground truth hard label in~\autoref{eq:loss}. The overall loss is then the linear combination of the distillation loss and the student loss.

\subsection{CRF distillation with node marginals}
Despite its popularity, knowledge distillation is generally only applicable to classification problems, as it is often intractable to calculate a distillation loss with respect to a structured learning model that involves exponentially large output label space. \newcite{kim2016sequence} conduct sequence-level knowledge distillation by training the student using the sequence-level predictions of the teacher. We argue that the approximation that only considers the best sequence-level predictions fails to transfer expressive information encoded in the teacher to the student.

To address the limitation, we instead propose an alternative strategy for knowledge distillation of a structured learning model, such as the CRF. Specifically, we train the student CRF by enforcing the token-level marginals produced by the teacher CRF. The marginal probability of a token $t$ being tagged as class $i$ is
\begin{equation}
p_{t,i} = p(y_t = i | \mathbf{x}),
\end{equation}
which can be efficiently computed by the forward-backward algorithm~\cite{rabiner1989tutorial}. The CRF distillation loss can be obtained by replacing the softmax probabilities in~\autoref{eq:loss} with the CRF node marginals.

We incorporate the proposed technique into \sysname~to distill knowledges of the pre-trained BiLSTM-CRF based NER models into a unified BiLSTM-CRF model. During training, we back-propagate the gradients of the CRF distillation loss with respect to not only the BiLSTM parameters but also the CRF transition parameters to optimize the student CRF model.

% ***********************************************************
% Experiment
\section{Experiments}
\label{sec:exp}

In this section, we compare \sysname~against competitive methods on three tasks related to NER with different tag sets: tag set extension, full tag set integration, and progressive NER. We derive datasets for the tasks based on two standard benchmark corpora on named entity recognition.

\begin{table} [ht!]
\centering
\small
%\addtolength{\tabcolsep}{-2pt}
\begin{tabular}{lrrr}
    \toprule
    Dataset &  Domain & \# of tokens & \# of entities \\ \midrule
     \multirow{6}{*}{OntoNotes-train} & NW  & 387,082 &  35,771 \\
     & BN & 180,300 & 17,573 \\
     & BC & 144,590 & 8,654 \\
     & MZ & 164,223 & 10,921 \\
     & TC & 81,144 & 2,233 \\
     & WB & 131,164 & 6,676  \\
      \cmidrule(l){2-4}
     & \text{Total} & 1,088,503 & 81,828 \\
     \cmidrule(l){2-4}
      \multirow{6}{*}{OntoNotes-dev} & NW & 52,618 & 4,883  \\
     & BN & 22,148 & 2,172 \\
     & BC & 26,550 & 1,459 \\
     & MZ & 15,422 & 1,232 \\
     & TC & 11,467 & 311 \\
      & WB & 19,519 & 1,009 \\
       \cmidrule(l){2-4}
     & \text{Total} & 147,724 & 11,066  \\
       \cmidrule(l){2-4}
      \multirow{6}{*}{OntoNotes-test} & NW & 49,235 & 4,696 \\
     & BN & 23,209 & 2,184 \\
     & BC &32,488 & 1,697  \\
     & MZ & 17,875 & 1,163 \\
     & TC & 10,976 & 380 \\
     & WB & 18,945 & 1,137 \\
      \cmidrule(l){2-4}
     & \text{Total} & 152,728 & 11,257  \\ \midrule
     I2B2-2014-train & Medical & 444,191 & 12,033  \\
     I2B2-2014-dev & Medical & 198,870 & 5,609  \\
     I2B2-2014-test & Medical & 414,661 & 11,591 \\
    \bottomrule
\end{tabular}
\caption{Dataset statistics. OnteNotes datasets are further divided based on the source domains of the documents: newswire (NW), broadcast news (BN)	, broadcast conversation (BC), magazine (MZ), telephone conversation (TC), and web data (WB).}
\label{tab:data}
\end{table}

\subsection{Data}
% I2B2 2014 and OntoNotes 5
In our experiments, we consider two standard NER datasets: OntoNotes 5.0~\cite{weischedel2013ontonotes} and I2B2 2014~\cite{stubbs2015annotating}. To our knowledge, they are by far the two largest annotated NER corpora for the general and medical domains respectively. The datasets are labeled with popular and diverse named entity types for their corresponding domains. In particular, OntoNotes contains $18$ entity types and I2B2'14 annotates $23$ entity types, among which only the \texttt{DATE} entity type aligns perfectly between the two tag sets. \autoref{tab:data} presents detailed statistics of the datasets. We use the OntoNotes train/development/test splits released for the CoNLL 2012 shared task.\footnote{Available at: \url{http://conll.cemantix.org/2012/data.html}} The original OntoNotes documents were drawn from a wide variety of sources, including newswire, broadcast news, broadcast conversation, magazine, telephone conversation, and web data.

% The i2b2 2014 dataset (Stubbs et al., 2015) was released as part of the 2014 i2b2/UTHealth shared task Track 1. It is the largest publicly available dataset for de-identification, which is a form of named-entity recognition where the entities are protected health information such as patients’ names and patients’ phone numbers.

\paragraph{Tag hierarchy}
We filter out I2B2'14 tags whose overall frequencies are less than $20$, as well as tags that conflict with the OntoNotes \texttt{CARDINAL} tag, which are not resolvable with the tag hierarchy. A total of $16$ I2B2'14 tags remained after the filtering.

We build the hierarchy similar to~\autoref{fig:tag-hierarchy}. In particular, the following hypernym-hyponym relationships between OntoNotes and I2B2'14 tag sets are included:
\begin{itemize} \itemsep -2pt
\footnotesize
\item \texttt{PERSON} : \{\texttt{DOCTOR}, \texttt{PATIENT}, \texttt{PERSON-OTHER}\}
\item \texttt{GPE} : \{\texttt{CITY}, \texttt{STATE}, \texttt{COUNTRY}\}
\item \texttt{ORG} : \{\texttt{HOSPITAL}, \texttt{ORGANIZATION}, \texttt{ORG-OTHER}\}
\end{itemize}
The rest of the tags are either OntoNotes-spec or I2B2'14-spec entity types, except for \texttt{DATA} that is shared by both tag sets.

\subsection{Experimental settings}
We train the models on the training datasets, tune the hyper-parameters on the development datasets, and evaluate the competitive systems on the test datasets. We report the standard evaluation metrics for NER: micro averaged precision, recall, and F1 score.

\paragraph{Competitive systems}
%We experiment with two neural network architectures for NER: BiLSTM and BiLSTM-CRF. BiLSTM replaces the CRF layer of BiLSTM-CRF with a linear softmax classifier layer for each token.
We consider four competitive NER methods that can work with multiple tag sets.  \textit{Post Processing} is a simple heuristic that resolves the conflicted tag predictions from different models by choosing the one with the highest (marginal) tag probability. \textit{Marginal CRF}~\cite{greenberg2018marginal,beryozkin2019joint} is the state-of-the-art (SOTA) approach for training a unified NER model with different tag sets. \textit{Neural Adapter}~\cite{chen2019transfer} is a transfer learning technique for sequence labeling that only uses source model and target data. It is the SOTA method for the progressive NER task. Finally, we compare the three variants of \sysname~that differ in the availability of the raw annotations. \textit{\sysname} solely uses the pre-trained models to perform knowledge distillation. \textit{\sysname-Data} also utilizes the labeled training data to produce the student loss when training the unified model. \textit{\sysname-Progressive} learns to transfer knowledge from a source model and a target dataset to a NER model. It is designed  particularly for the progressive NER problem.

One advantage of \sysname~over Marginal CRF is that it is comparable with both local and global probabilistic models. Thus, we are able to include results for both BiLSTM-based NER systems and BiLSTM-CRF-based systems.

\paragraph{Parameter tuning}
We follow~\newcite{yang2018design} to set the hyper-parameter values of the BiLSTM-CRF models used in all the experiments. Specifically, we set the batch size to $10$, the learning rate to $0.015$, and the dropout rate to $0.5$. We adopt one-layer word-level BiLSTM model and one-layer character-level uni-directional LSTM model, as we found it performs slightly better than the BiLSTM model for capturing the character-level information. The sizes of the LSTM hidden state vectors are $50$ and $200$ for the character-level and word-level models respectively. The sizes of the input embedding vectors are set as $30$ for the character-level LSTM model and $100$ for the word-level LSTM model, as we used the pre-trained $100$-dimensional GloVe vectors to initialize the word embedding matrix. The linear combination coefficient $\alpha$ for combining the distillation loss and the student loss was chose from $\{ 0.2, 0.4, 0.6, 0.8 \}$.

\subsection{Tag set extension}
\label{sec:exp:tag-ext}

\begin{table} [ht!]
\centering
\small
\addtolength{\tabcolsep}{-2pt}
\begin{tabular}{lcccccc}
    \toprule
    System & BN & BC & WB & MZ & TC  & AVG \\ \midrule
     \multicolumn{7}{l}{\it BiLSTM systems} \\
     Post Processing & 80.0 & 50.8 & 65.3 & 69.7 & 62.3 & 65.6 \\
     \sysname            & 83.9 & 73.7 & 70.7 & 72.9 & 65.6 & 73.4 \\
     \sysname-Data   & 85.1 & 74.3 & 72.7 & 74.7 & 67.8 & 74.6 \\ [5pt]
     \multicolumn{7}{l}{\it BiLSTM-CRF systems} \\
     Post Processing & 82.6 & 67.2 & 67.3 & 70.1 & 66.3 & 70.7 \\
     Marginal CRF     & \textbf{87.2} & \textbf{78.2} & 73.8 & 76.7 & 69.2 & 77.0 \\
     \sysname            & 86.6 & 76.5 & 73.3 & 75.7 & 69.5 & 76.3 \\
     \sysname-Data   & 86.9 & 77.7 & \textbf{75.1} & \textbf{76.8} & \textbf{70.2} & \textbf{77.3} \\
       \bottomrule
\end{tabular}
\caption{F1 score results for tag set extension experiments. The best results are in \textbf{bold}.}
\label{tab:res:tag-ext}
\end{table}

Our first set of experiments are motivated by the real world scenario related to selective annotation discussed in~\autoref{sec:intro}.  When building a NER dataset for a special domain, to avoid expensive annotations, we can selectively annotate entities that are specifically for that domain, and rely on general NER datasets to handle the rest of named entities. We treat OntoNotes 5.0 as the testbed for the tag set extension experiments, as it contains documents from multiple source domains.

To mimic the scenario, we treat the newswire (NW) domain as the general domain and each of the other OntoNotes domains as specific target domains. The source tag set covers the five most frequent entity types in the newswire domain that are \texttt{ORG}, \texttt{DATE}, \texttt{GPE}, \texttt{PERSON}, and \texttt{CARDINAL}, and the target tag set consists of the rest of OntoNotes entity types. The source annotations corresponding to the target tag set and the target annotations corresponding to the source tag set are wiped out --- we simply retag the relevant tokens to \texttt{O}.

For each extension setting, we train on the training sets with the corresponding source and target tag sets, and tune hyper-parameters and evaluate on the development and test sets of the target domain with the original full tag set. The results are shown in~\autoref{tab:res:tag-ext}, indicating that \sysname~performs generally on par with Marginal CRF when the training datasets are given. When \sysname~has access only to the pre-trained NER models, Marginal CRF slightly outperforms it. However, \sysname~provides dramatic flexibility on the required form of NER resources. Both Marginal CRF and \sysname-based systems considerably outperform the Post Processing baseline, showing the power of joint training a NER model over multiple tag sets. Finally, knowledge distillation of CRF models yields much more superior results than distillation of classification models on the NER task, which suggests that \sysname~is indeed able to properly distill knowledge from a CRF teacher into a CRF student.

\subsection{Full tag set integration}

\begin{table} [ht!]
\centering
\small
\addtolength{\tabcolsep}{-2pt}
\begin{tabular}{lcccccc}
    \toprule
     \multirow{2}{*}{System} & \multicolumn{3}{c}{OntoNotes 5.0} & \multicolumn{3}{c}{I2B2 2014} \\
    \cmidrule(l){2-4} \cmidrule(l){5-7}
          & Prec & Rec & F1 & Prec & Rec & F1 \\ \midrule
     \multicolumn{7}{l}{\it BiLSTM systems} \\
     Post Processing & 81.3 & 80.9 & 81.1 & 90.7 & 84.0 & 87.3 \\
     \sysname            & 83.1 & 86.5 & 84.8 & 87.9 & 86.4 & 87.2 \\
     \sysname-Data   & 84.0 & 86.6 & 85.3 & 87.9 & 87.2 & 87.5 \\ [5pt]
     \multicolumn{7}{l}{\it BiLSTM-CRF systems} \\
     Post Processing & 85.6 & 83.1 & 84.3 & 90.6 & 80.7 & 85.3 \\
     Marginal CRF     & 87.4 & \textbf{87.3} & \textbf{87.3} & \textbf{92.6} & \textbf{88.0} & \textbf{90.2} \\
     \sysname            & 87.2 & 87.2 & 87.2 & 91.9 & 87.8 & 89.8 \\
     \sysname-Data   & \textbf{87.5} & 87.1 & \textbf{87.3} & 92.2 & 87.9 & 90.0 \\
       \bottomrule
\end{tabular}
\caption{Precision, Recall, and F1 score results for full tag set integration experiments. The best results are in \textbf{bold}.}
\label{tab:res:tag-int}
\end{table}

Following~\newcite{beryozkin2019joint}, we also experiment with the full tag set integration evaluation setting. We need to consolidate the heterogeneous tag sets of OntoNotes and I2B2'14 to have the best of both worlds. In particular, we train a named entity tagger on the training sets of the OntoNotes and I2B2'14 corpora with the unified tag set. Compared to the setting used by~\newcite{beryozkin2019joint}, where they integrate the tag sets of I2B2'06 and I2B2'14, we argue that our setting is more realistic and also more challenging. The I2B2'06 tag set is basically a subset of the I2B2'14 tag set that generally regard similar entity types. As we have shown before, the tag sets of OntoNotes and I2B2'14 focus on very different entity types, which provide a better platform to evaluate the systems on consolidating heterogeneous tag sets.

The full tag set integration results are presented in~\autoref{tab:res:tag-int}. As shown, both \sysname~and \sysname-Data achieve similar results as Marginal CRF in the full tag set integration setting. Again, joint-training-based methods significantly outperform the Post Processing baseline for both BiLSTM-based and BiLSTM-CRF-based systems. The results show that \sysname~can principally integrate NER models trained with heterogeneous tag sets using an easy-to-build tag hierarchy.

\subsection{Progressive NER}

\begin{table} [ht!]
\centering
\small
\addtolength{\tabcolsep}{-2pt}
\begin{tabular}{lcccccc}
    \toprule
    System & BN & BC & MZ & TC & WB  & AVG \\ \midrule
     \multicolumn{7}{l}{\it Unlabeled source data is available} \\
     \sysname-Progressive & \textbf{86.6} & \textbf{77.3} & \textbf{74.4} & \textbf{76.8} & \textbf{69.0} & \textbf{76.8} \\ [5pt]
     \multicolumn{7}{l}{\it Unlabeled source data is not available} \\
     Post Processing & 82.6 & 67.2 & 67.3 & 70.1 & 66.3 & 70.7 \\
     Neural Adapter & 70.5 & 61.1 & 59.8 & 60.1 & 63.1 & 62.9 \\
     \sysname-Progressive & 84.3 & 69.7 & 68.2 & 71.4 & 67.8 & 72.3 \\
       \bottomrule
\end{tabular}
\caption{F1 score results for progressive NER experiments. The best results are in \textbf{bold}.}
\label{tab:res:tag-pro}
\end{table}

To demonstrate \sysname's flexibility on the existence of source data, we consider the progressive NER setting proposed by~\newcite{chen2019transfer}. Specifically, we conduct another set of experiments that are similar to the tag set extension experiments described in~\autoref{sec:exp:tag-ext}. The key distinction is that we assume that only the source model and the target training data are available in the progressive NER setting. As \sysname~needs to access unlabeled data to perform knowledge distillation, we further experiment with two variants of \sysname. In the first variant, we employ unlabeled data from both the source and target domains to train the distiller. In the second variant, we assume that the source unlabeled data is not available and distill both the source and target models using the unlabeled data from the target domain. The variant erects a fairer comparison with Neural Adapter. %we initialize the parameters of the unified NER model with the source model parameters and then distill the NER model trained on the target data into the unified model.

As shown in~\autoref{tab:res:tag-pro}, \sysname~leads to a significant boost in performance compared to the SOTA method Neural Adapter. Neural Adapter~\cite{chen2019transfer} initializes the parameters of the unified NER model with the source model parameters and then fine-tune the model on the target data. Collisions rooted from different tag sets still occur during the fine-tuning, resulting in less optimal transfer performance. Among the two \sysname~variants, the one with access to unlabeled source data performs better, suggesting in-domain unlabeled data is crucial for effective knowledge distillation.

% Our \sysname~variant  unlabeled source data is not available

% collision while \sysname resolve the collisions during fine tuning, thus gives rise to better results.

% Evaluation 1: original annotations are available
% - disjoint tag sets
% - heterogeneous tag sets
% Evaluation 2: partial or all annotations are not available
% - progressive NER
% - where to sample the data

% ***********************************************************
% Related Work
\section{Related Work}
\label{sec:related}

\paragraph{Neural NER}
Early works on NER focused on engineering useful linguistic features for the task~\cite{tjong2003introduction,nadeau2007survey}. With the resurgence of neural networks, recent works have been focused on adopting neural models for the task of NER.
%Compared to traditional approaches, neural methods are capable of learning expressive representations that are useful for the specific task, which avoid expensive feature engineering.
\newcite{lample2016neural} proposed a BiLSTM-CRF model that consists of a BiLSTM layer to learn feature representations from the input and a CRF layer to model the interdependencies between adjacent labels.  A similar neural architecture for NER was introduced by~\newcite{ma2016end}, where the subword unit information was modeled with character-level CNNs instead of the BiLSTM networks used in BiLSTM-CRF.  In this work, we use BiLSTM-CRF as our base model, as it is widely used in literature. However, \sysname~works with any probabilistic model in principle.

\paragraph{NER with multiple tag sets}
Early research efforts made on NER with multiple tag sets are mostly related to NER with partially annotated data. \newcite{bellare2007learning} proposed a missing label linear-chain CRF that was essentially a latent variable CRF~\cite{quattoni2005conditional} for a set of NLP tasks with partially annotated data. \newcite{greenberg2018marginal} presented a marginal CRF model that was a variant of the latent variable CRF method. Marginal CRF achieved promising performance in joint training a biomedical NER model with multiple tag sets. The method was extended by~\cite{beryozkin2019joint} to handle datasets with heterogeneous label sets. Marginal CRF needs to be trained on the data annotations, while \sysname~can be learned from pre-trained models with different tag sets.

\paragraph{Progressive learning}
Standard machine learning methods assume the data for training and testing has the same feature space and distribution (i.e., the independent, identically distributed assumption). In recent years, the technique of transfer learning has emerged to relax this assumption that makes machine learning models more applicable to real world applications. Progressive learning falls in a more specific transfer learning category, in which we need to transfer knowledge from a source model using a target dataset that involves additional labels. \newcite{venkatesan2016novel} adopted progressive learning techniques to a set of multi-class classification problems. In particular, they remodeled a single layer feed-forward network by increasing the number of new neurons and interconnections while encountering unseen class labels in the dataset.
The technique was then further extended to NER by~\newcite{chen2019transfer}, where they copied parameters of the source model to a new model with enlarged label space and then fine tuned the model on the target data. Progressive NER is a special case of the NER model unification problem discussed in the work. We show that \sysname~can successfully solve a set of related tasks including progressive NER.

% ***********************************************************
% Conclusion and Future Work
\section{Conclusion}
\label{sec:con}
In this paper, we present \sysname, a knowledge distillation based approach to unify NER models pre-trained on different tag sets into a centralized model. It is capable of effectively distilling knowledge from CRF teachers into a CRF student. \sysname~can work with heterogeneous NER resources in the form of either a dataset or a model, making it a flexible framework to consolidate NER systems across different domains. We conduct extensive experiments on three related tasks. Despite the unavailability of annotations, \sysname~performs on par with the state-of-the-art (SOAT) method on joint training of NER models with either disjoint tag sets or heterogeneous tag sets. It significantly outperforms the SOAT model on the progressive NER task. In the future, we would like to extend \sysname~to handle tag sets that involve partially overlapping tags. We are also interested in applying the CRF distillation technique to other NLP problems such as part-of-speech tagging and chunking.

% ***********************************************************
% Acknowledgments
%\input{ack}

% ***********************************************************
\bibliographystyle{acl_natbib}
\bibliography{cite-strings,cites,cite-definitions}

\begin{thebibliography}{21}
\expandafter\ifx\csname natexlab\endcsname\relax\def\natexlab#1{#1}\fi

\bibitem[{Bellare and McCallum(2007)}]{bellare2007learning}
Kedar Bellare and Andrew McCallum. 2007.
\newblock Learning extractors from unlabeled text using relevant databases.
\newblock In \emph{Sixth international workshop on information integration on
  the web}.

\bibitem[{Beryozkin et~al.(2019)Beryozkin, Drori, Gilon, Hartman, and
  Szpektor}]{beryozkin2019joint}
Genady Beryozkin, Yoel Drori, Oren Gilon, Tzvika Hartman, and Idan Szpektor.
  2019.
\newblock A joint named-entity recognizer for heterogeneous tag-setsusing a tag
  hierarchy.
\newblock In \emph{{Proceedings of the Association for Computational
  Linguistics (ACL)}}.

\bibitem[{Buciluǎ et~al.(2006)Buciluǎ, Caruana, and
  Niculescu-Mizil}]{bucilua2006model}
Cristian Buciluǎ, Rich Caruana, and Alexandru Niculescu-Mizil. 2006.
\newblock Model compression.
\newblock In \emph{{Proceedings of Knowledge Discovery and Data Mining (KDD)}}.

\bibitem[{Chen and Moschitti(2019)}]{chen2019transfer}
Lingzhen Chen and Alessandro Moschitti. 2019.
\newblock Transfer learning for sequence labeling using source model and target
  data.
\newblock In \emph{{Proceedings of the National Conference on Artificial
  Intelligence (AAAI)}}.

\bibitem[{Greenberg et~al.(2018)Greenberg, Bansal, Verga, and
  McCallum}]{greenberg2018marginal}
Nathan Greenberg, Trapit Bansal, Patrick Verga, and Andrew McCallum. 2018.
\newblock Marginal likelihood training of bilstm-crf for biomedical named
  entity recognition from disjoint label sets.
\newblock In \emph{{Proceedings of Empirical Methods for Natural Language
  Processing (EMNLP)}}.

\bibitem[{Hinton et~al.(2015)Hinton, Vinyals, and Dean}]{hinton2015distilling}
Geoffrey Hinton, Oriol Vinyals, and Jeff Dean. 2015.
\newblock Distilling the knowledge in a neural network.
\newblock \emph{arXiv preprint arXiv:1503.02531}.

\bibitem[{Kim and Rush(2016)}]{kim2016sequence}
Yoon Kim and Alexander~M Rush. 2016.
\newblock Sequence-level knowledge distillation.
\newblock In \emph{{Proceedings of Empirical Methods for Natural Language
  Processing (EMNLP)}}.

\bibitem[{Lafferty et~al.(2001)Lafferty, McCallum, and
  Pereira}]{lafferty2001conditional}
John Lafferty, Andrew McCallum, and Fernando~CN Pereira. 2001.
\newblock Conditional random fields: Probabilistic models for segmenting and
  labeling sequence data.
\newblock In \emph{{Proceedings of the International Conference on Machine
  Learning (ICML)}}.

\bibitem[{Lample et~al.(2016)Lample, Ballesteros, Subramanian, Kawakami, and
  Dyer}]{lample2016neural}
Guillaume Lample, Miguel Ballesteros, Sandeep Subramanian, Kazuya Kawakami, and
  Chris Dyer. 2016.
\newblock Neural architectures for named entity recognition.
\newblock In \emph{{Proceedings of the North American Chapter of the
  Association for Computational Linguistics (NAACL)}}.

\bibitem[{Ma and Hovy(2016)}]{ma2016end}
Xuezhe Ma and Eduard Hovy. 2016.
\newblock End-to-end sequence labeling via bi-directional lstm-cnns-crf.
\newblock In \emph{{Proceedings of the Association for Computational
  Linguistics (ACL)}}.

\bibitem[{Mintz et~al.(2009)Mintz, Bills, Snow, and
  Jurafsky}]{mintz2009distant}
Mike Mintz, Steven Bills, Rion Snow, and Dan Jurafsky. 2009.
\newblock Distant supervision for relation extraction without labeled data.
\newblock In \emph{{Proceedings of the Association for Computational
  Linguistics (ACL)}}.

\bibitem[{Nadeau and Sekine(2007)}]{nadeau2007survey}
David Nadeau and Satoshi Sekine. 2007.
\newblock A survey of named entity recognition and classification.
\newblock \emph{Lingvisticae Investigationes}.

\bibitem[{Pennington et~al.(2014)Pennington, Socher, and
  Manning}]{pennington2014glove}
Jeffrey Pennington, Richard Socher, and Christopher Manning. 2014.
\newblock Glove: Global vectors for word representation.
\newblock In \emph{{Proceedings of Empirical Methods for Natural Language
  Processing (EMNLP)}}.

\bibitem[{Quattoni et~al.(2005)Quattoni, Collins, and
  Darrell}]{quattoni2005conditional}
Ariadna Quattoni, Michael Collins, and Trevor Darrell. 2005.
\newblock Conditional random fields for object recognition.
\newblock In \emph{Advances in neural information processing systems}.

\bibitem[{Rabiner(1989)}]{rabiner1989tutorial}
Lawrence~R Rabiner. 1989.
\newblock A tutorial on hidden markov models and selected applications in
  speech recognition.
\newblock \emph{Proceedings of the IEEE}.

\bibitem[{Stubbs and Uzuner(2015)}]{stubbs2015annotating}
Amber Stubbs and {\"O}zlem Uzuner. 2015.
\newblock Annotating longitudinal clinical narratives for de-identification:
  The 2014 i2b2/uthealth corpus.
\newblock \emph{Journal of biomedical informatics}.

\bibitem[{Tjong Kim~Sang and De~Meulder(2003)}]{tjong2003introduction}
Erik~F Tjong Kim~Sang and Fien De~Meulder. 2003.
\newblock Introduction to the conll-2003 shared task: language-independent
  named entity recognition.
\newblock In \emph{Proceedings of the seventh conference on Natural language
  learning at HLT-NAACL 2003-Volume 4}.

\bibitem[{Venkatesan and Er(2016)}]{venkatesan2016novel}
Rajasekar Venkatesan and Meng~Joo Er. 2016.
\newblock A novel progressive learning technique for multi-class
  classification.
\newblock \emph{Neurocomputing}.

\bibitem[{Weischedel et~al.(2013)Weischedel, Palmer, Marcus, Hovy, Pradhan,
  Ramshaw, Xue, Taylor, Kaufman, Franchini et~al.}]{weischedel2013ontonotes}
Ralph Weischedel, Martha Palmer, Mitchell Marcus, Eduard Hovy, Sameer Pradhan,
  Lance Ramshaw, Nianwen Xue, Ann Taylor, Jeff Kaufman, Michelle Franchini,
  et~al. 2013.
\newblock Ontonotes release 5.0 ldc2013t19.
\newblock \emph{Linguistic Data Consortium, Philadelphia, PA}.

\bibitem[{Yang et~al.(2018)Yang, Liang, and Zhang}]{yang2018design}
Jie Yang, Shuailong Liang, and Yue Zhang. 2018.
\newblock Design challenges and misconceptions in neural sequence labeling.
\newblock In \emph{Proceedings of the 27th International Conference on
  Computational Linguistics (COLING)}.

\bibitem[{Yih et~al.(2015)Yih, Chang, He, and Gao}]{yih2015semantic}
Scott Wen-tau Yih, Ming-Wei Chang, Xiaodong He, and Jianfeng Gao. 2015.
\newblock Semantic parsing via staged query graph generation: Question
  answering with knowledge base.
\newblock In \emph{{Proceedings of the Association for Computational
  Linguistics (ACL)}}.

\end{thebibliography}

\end{document}